\documentclass[conference]{IEEEtran}
\IEEEoverridecommandlockouts
\usepackage{amsmath,amssymb,amsfonts}
\usepackage{graphicx}
\usepackage{xcolor}
\usepackage{float}
\usepackage{tabularx}
\usepackage{multirow} 
\usepackage{hyperref}
\usepackage{booktabs}

\def\BibTeX{{\rm B\kern-.05em{\sc i\kern-.025em b}\kern-.08em
    T\kern-.1667em\lower.7ex\hbox{E}\kern-.125emX}}
\begin{document}

\title{Development of Real-time Rendering Technology for High-Precision Models in Autonomous Driving\\}

\author{\IEEEauthorblockN{Wencheng ZHANG}
\and
\IEEEauthorblockN{Chengyi WANG}}

\maketitle

\begin{abstract}
Our autonomous driving simulation lab produces a high-precision 3D model simulating the parking lot. 
However, the current model still has poor rendering quality in some aspects. 
In this work, we develop a system to improve the rendering of the model and evaluate the quality of the rendered model.
\end{abstract}

\section{Introduction}
Computer graphics rendering technology is essential for autonomous driving simulation. Specifically, when engineers attempt to run a simulation, they must first create a 3D model that specifies how geometries should be placed in the virtual world \cite{lan2016development1,lan2016development2}. 
Then it comes to rendering. ``Texture", including color, roughness, and so on, should be specified for each geometry. Only after being rendered, the 3D model can simulate the real world with a lifelike look. 
After that, engineers can test their autonomous driving algorithms in the model.

In this work, we already have a 3D model of the parking lot of our campus, but now it only has a low-quality rendering. 
Our task is to improve the rendering quality and use methods to assess the quality scientifically.

Our project is of great importance if it's completed well because our lab's engineers are going to test autonomous driving algorithms using this parking lot model.
Without a high-quality rendering, the training process is pointless.

\section{Related Work}
\subsection{Introduction to Image Quality Assessment}
We are using the Image Quality Assessment (IQA) algorithms to scientifically assess the rendering quality.
The target of IQA is to quantify the visual distortion and produce the perceptive quality score of the image. It is often applied in fields such as image acquisition, transmission, compression, restoration, and enhancement.
IQA is often classified into three types, Full-referenced (FR), Reduced-referenced (RR), and Non-referenced (NR), which differ in their level of reliance on a reference image. 
Specifically, FR-IQA assesses with the help of a high-quality reference image that is pixel-aligned with the test image, RR-IQA uses parts of the reference's information, such as some features of the reference and NR-IQA is not using any reference. 

\subsection{Some featured IQA Researches}
A method known as CVRKD-IQA proposed by Yin et al. \cite{CVRKD_2022} is especially suitable for our underground parking lot rendering research. 
Knowledge distillation is used for feature extraction from pictures and training the NAR-student agent \cite{liu2022towards,lan2022semantic}.
Under the verification of various data sets and algorithms, the results of the IQA algorithm have sufficient credibility.
Bosse et al. \cite{WaDIQaM_2018} implemented an FR-IQA and NR-IQA method by Deep Neural Networks named WaDIQaM.
The pre-study and training of different feature fusion strategies make the DNN system effective on FR and NR IQA scenes.

Hossein and Peyman \cite{NIMA_2018} introduced a neural image assessment algorithm trained on both aesthetic and pixel-level quality datasets. 
This CNN-based NR-IQA method is called NIMA. 
NIMA is a pre-trained IQA algorithm and is popular for effectively predicting the distribution of quality ratings, rather than just the mean scores.

\subsection{The Practice of IQA in Rendering}
Julian et al. \cite{amann2013using} propose their ideas about testing rendering. 
As game engines and renderers have increasingly high complexity, they use IQA to have an automatic test on the software, that is testing whether the rendering result is right. 
Eventually, a ``right" or ``wrong" result will be given for each test.

In some other studies \cite{lavou2014a}, it's proposed that most applications of computer rendering methods require perceptually plausible solutions rather than physically accurate results. We can use some perceptual metrics to compute the visible difference between the test image and the reference image.

Here are some examples of the IQA used by rendering experts.
HDRVDP algorithm uses FR-IQA to evaluate rendering \cite{mantiuk2011a}. 
NoRM algorithm uses NR-IQA to find artefacts in rendering \cite{herzog2012a}.

In our work, we are trying to compare our model to scenes in real life as we can easily take pictures of our university's parking lot. 
We want to quantify the similarity of our model to the real parking lot and use the result of this metric to help us improve the model.

\section{Proposed Approach}

The proposed approach and plan for the project are as below:

\begin{enumerate}
\item Find appropriate algorithms to scientifically assess rendering quality.
\item Find out all the conspicuous and widespread flaws in the 3D model by eyes, do an analysis of their cause, and start improving the model.
\item After improving the model, we use IQA algorithms to verify the improvement.
\item Eventually, we train the autonomous vehicles in the new parking lot model and see if the training result will be improved because of the improved rendering.
\end{enumerate}

The progress we have made is as below:
\begin{enumerate}
\item The 3D parking lot model is successfully run in Unreal Engine 4 and we have made progress in improving rendering.
\item We have already chosen some assessment algorithms, and successfully use them to assess rendering in the 3D parking lot model.
\end{enumerate}

\section{Improvement in Rendering}
We tried our preliminary improvement in the rendering of the parking lot model, including using global illumination and materials modification \cite{lan2022vision}.
Here's a picture showing the effect of adding global illumination and modifying materials.

\begin{figure}[!ht]
\centering
\includegraphics[width=0.3\textwidth]{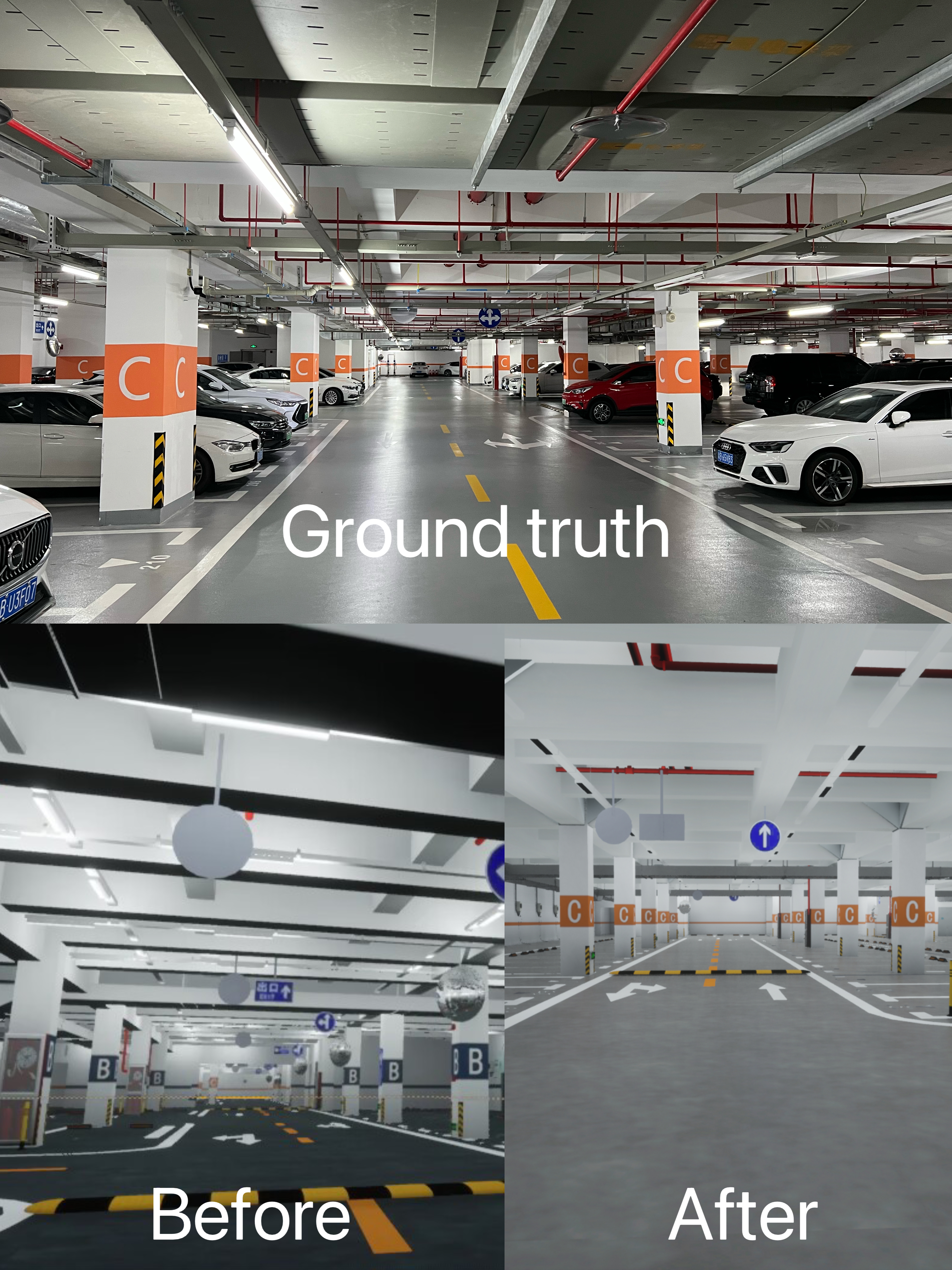}
\caption{Improvement in rendering.}
\end{figure} 
\label{fig:rendering-upgrade}

\subsection{Global Illumination}

Illumination can be categorized into two types \cite{lan2022class,gao2021neat}, one is direct illumination, and the other is indirect illumination (the same concept as global illumination). An object can be seen by our eyes only if it perceives light and bounces the light toward our eyes. If the light bounces twice or more before being perceived by us, it is called indirect illumination. Specifically, direct illumination is easy to be rendered in a game engine, but indirect illumination is not.
The effect of global illumination is huge, especially when some objects cannot perceive direct light \cite{GI}.

\begin{figure}[!ht]
\centering
\includegraphics[width=0.3\textwidth]{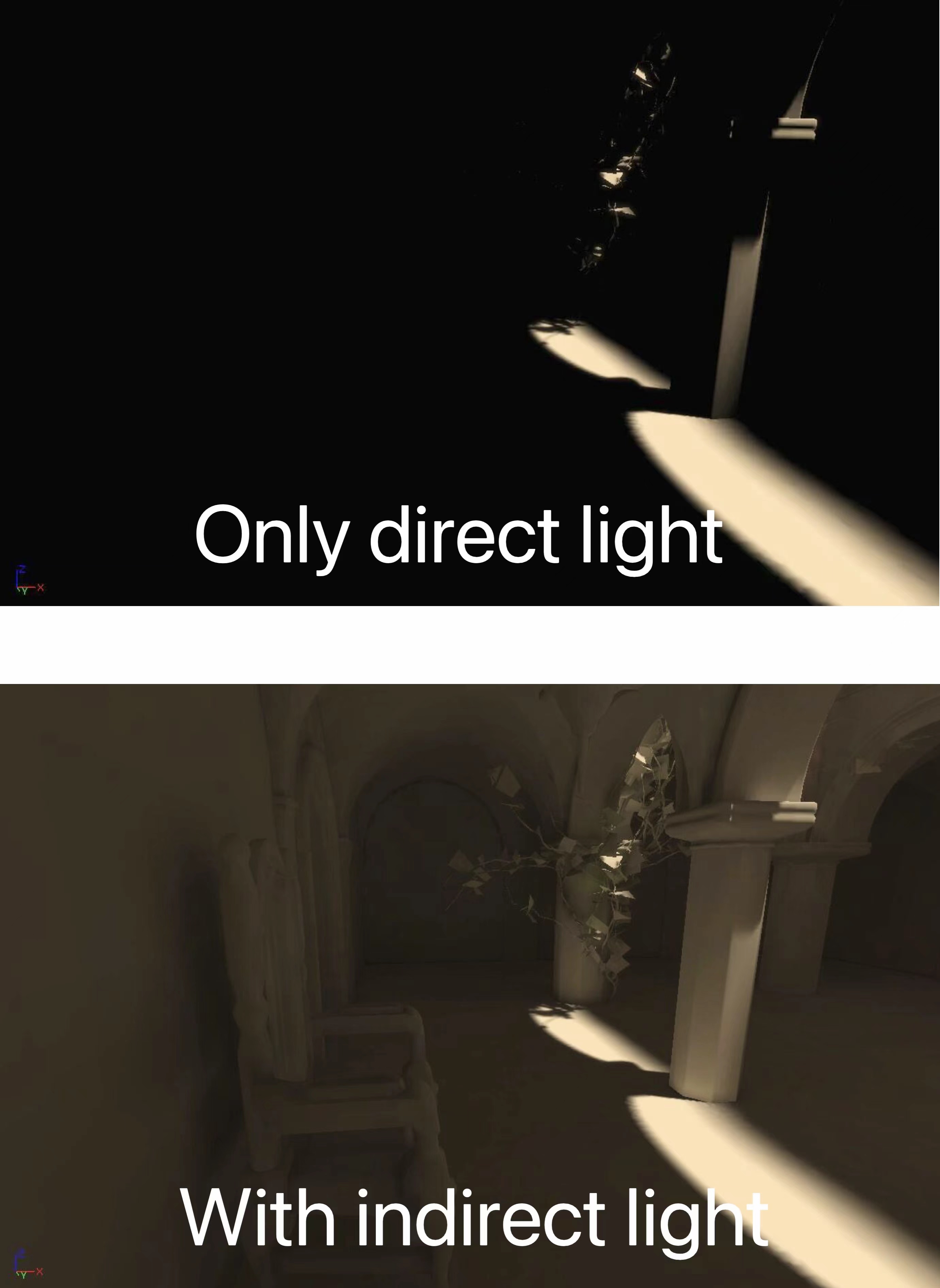}
\caption{ Global illumination's effect }
\end{figure} 
\label{fig:gi-effect}

The industry comes up with many algorithms to implement global illumination.

Our working tool, Unreal Engine 4, provides two ways for users, Precomputed Lighting and Dynamic Lighting \cite{epic_games_2020}.

In our parking lot model, nothing moves. 
Therefore we should utilize precomputed lighting, which can bring us high-quality global illumination and high running speed \cite{luksch_wimmer_schwarzler_2019}.

Here’s an image showing the model’s appearance with and without precomputed GI (Global Illumination). 

\begin{figure}[!ht]
\centering
\includegraphics[width=0.3\textwidth]{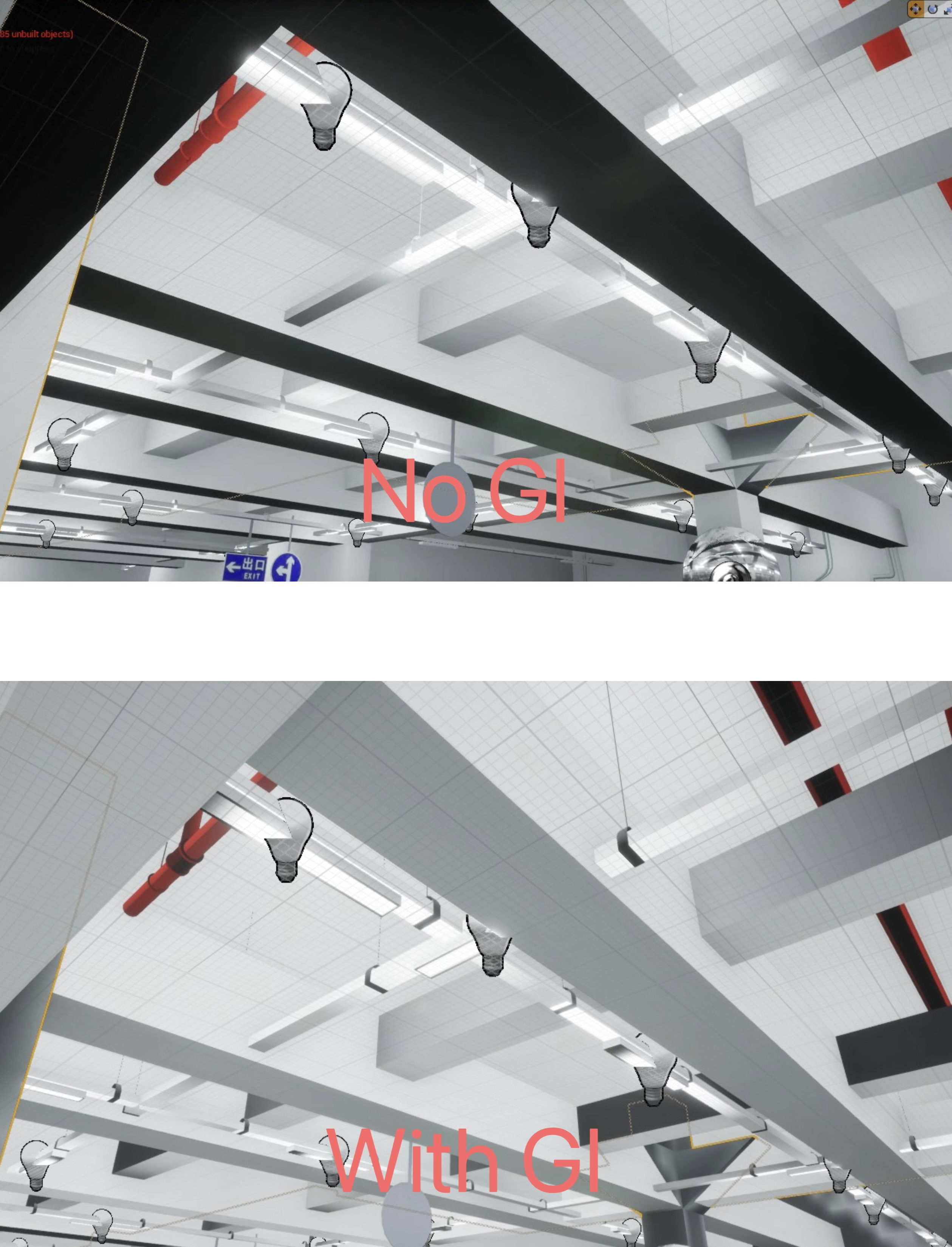}
\caption{The effect of having Global illumination in our model.}
\end{figure} 
\label{fig:model-comparison}

\subsection{Material Modification}
A 3D model is first made up of a lot of polygons, which we called a polygon mesh. 
A mesh only has a geometric shape but no color.
To emulate the world better, the mesh needs color \cite{botsch_2011}. 
Materials are assets to control the appearance of polygon meshes.

Making the right materials is a central topic in Computer Graphics, the key to making our model look like the real one. At the first glance at the original model, we can spot some objects that have the wrong materials. We use APIs provided by our tool, Unreal Engine 4, to modify the attributes of the materials, such as color, roughness \cite{DisneyBRDF}.
Here I put images showing two adjustments I did to the materials of the floor and the doors.

The floor now has a lighter gray color and a more shiny look that shows the reflection of the light. The doors get rid of the reflection of the light and the color is also adjusted.

\begin{figure}[!ht]
\centering
\includegraphics[width=0.3\textwidth]{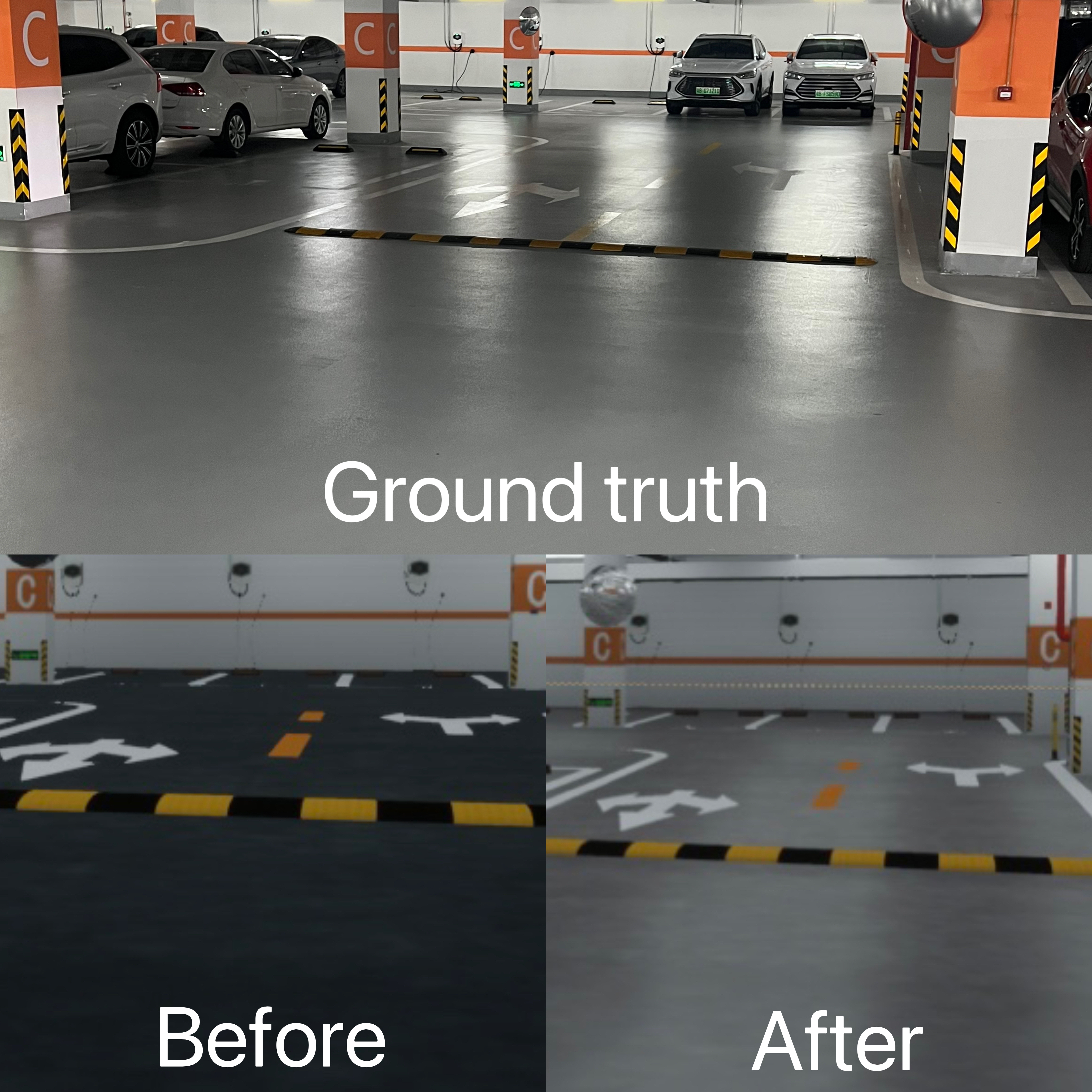}
\caption{Adjustment of the floor.}
\end{figure} 
\label{fig:adjust-floor}

\begin{figure}[!ht] \centering
\includegraphics[width=0.3\textwidth]{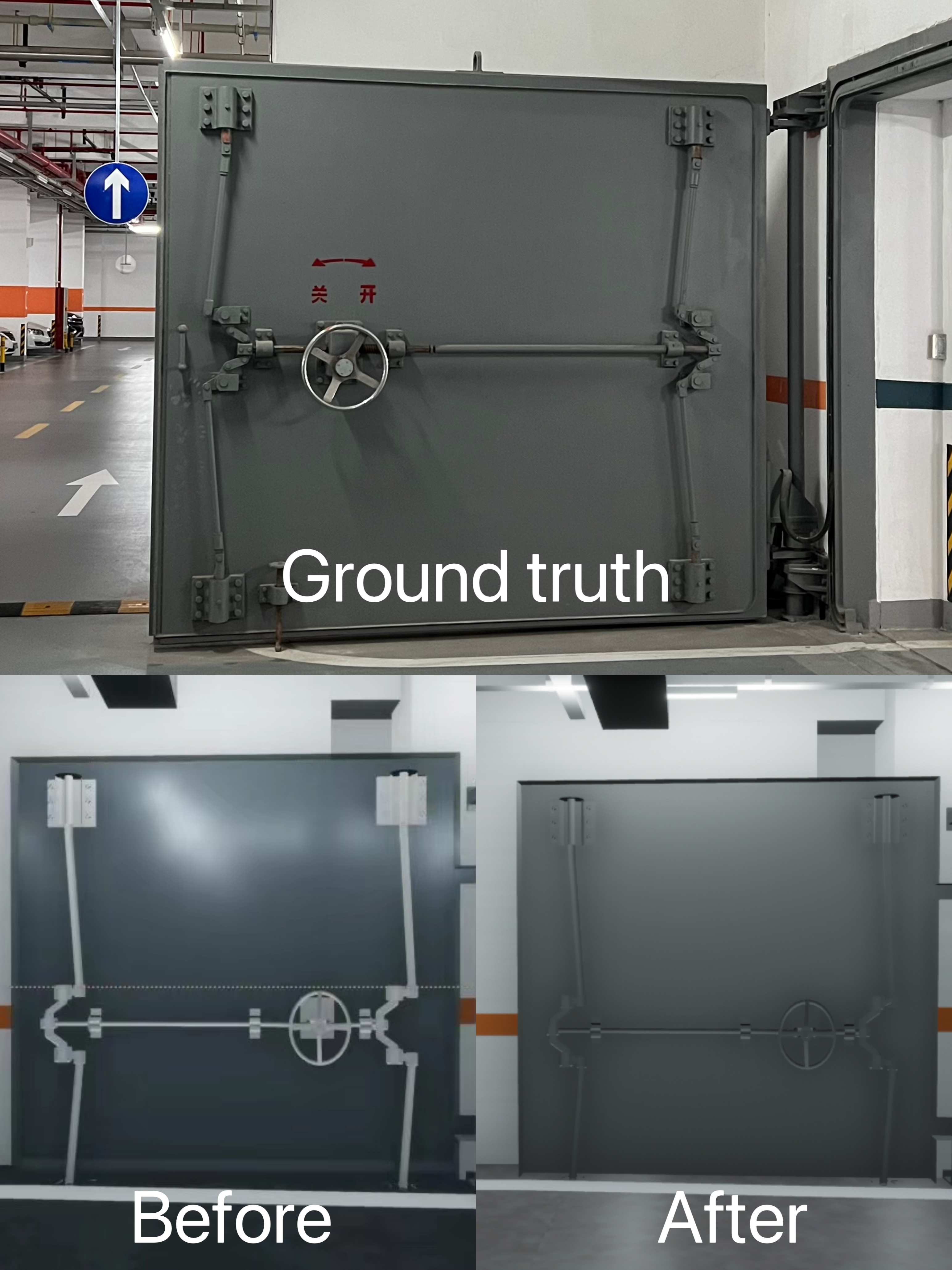}
\caption{Adjustment of the doors.}
\end{figure} 
\label{fig:adjust-door}

\section{Evaluation of the Project}
As we have already found some suitable IQA algorithms, which are CVRKD-IQA, WaDIQaM, and NIMA, we now use the IQA algorithms to have a test on the following three sets of image comparison and see if the scores are improved which can verify the improvement of rendering. 
The test results are shown in \autoref{tab:iqa}.

\begin{figure}[!ht] \centering
\includegraphics[width=0.3\textwidth]{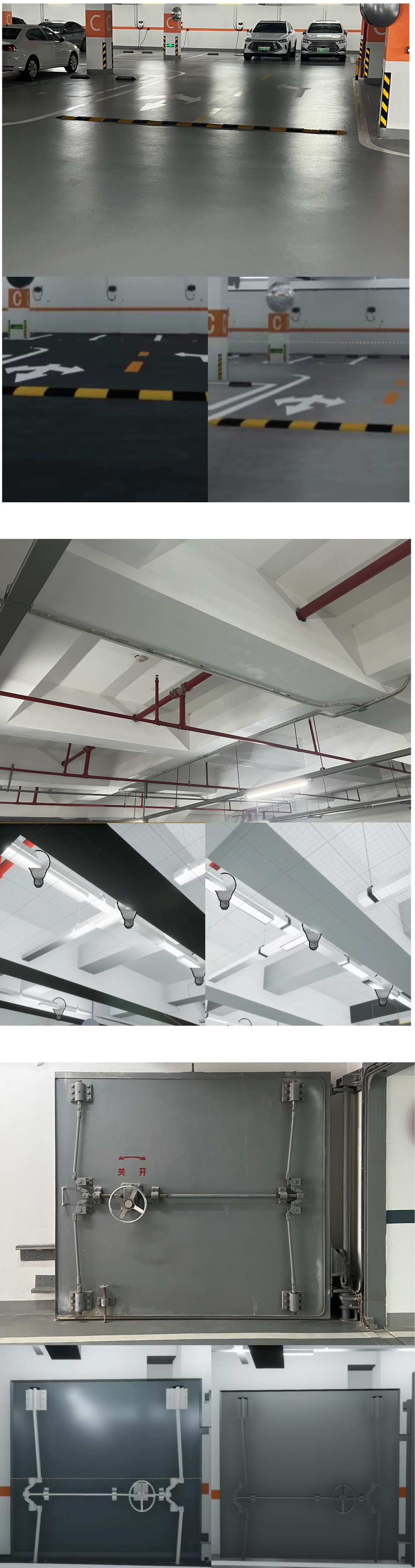}
\caption{ Three sets of comparison. For each set, the upper image is for ground truth; the bottom left is for the model before improvement; the bottom right is for the model after improvement}
\label{fig:test-scenes}
\end{figure}

\begin{table}[!ht] \centering
    \setlength\tabcolsep{4pt}
    \caption{IQA results}
    \begin{tabular}{|l|c c c|c c c|} 
      \hline
      \multirow{2}{*}{\textbf{Algorithm}} & \multicolumn{3}{c|}{\textbf{Original Rendering}} & \multicolumn{3}{c|}{\textbf{Improved Rendering}}\\
      \cline{2-7}
        & \textbf{Scene1} & \textbf{Scene2} & \textbf{Scene3} & \textbf{Scene1} & \textbf{Scene2} & \textbf{Scene3}\\
      \hline
      CVRKD & -1.1654 & 0.9932 & -0.3112 & -1.0244 & 1.5823 & -0.0749 \\
      WaDIQaM & -0.4971 & 0.3884 & -0.1692 & -0.4608 & 0.5792 & 0.1596\\
      NIMA & 0.4080 & 0.0202 & -0.6083 & 0.4673 & 0.1761 & -0.4622\\
      \hline
    \end{tabular} \label{tab:iqa}
\end{table}

The results show that our model after modification performs better in these image quality assessment systems. Although due to the limited number of selected assessment algorithms, the verification is still not universal enough to prove that our model after modification is strictly better than the original model, at least we have reasons to believe that our modification has effectively improved the rendering quality of Sustech's COE underground parking lot model. 

The results given by the three IQAs show that the improvement due to the modification in illumination is very stable on both FR and NR IQAs.
However, though the modification of material has positive impacts on the results given by the three IQAs, the impacts are unbalanced among the three.
Taking CVRKD as an example, CVRKD's evaluation of image quality is divided into two parts-comparing the test image with the true value and quality evaluation according to the results of data set training. Changing the material brings the test image closer to the ground truth, resulting in a great improvement in grades \cite{CVRKD_2022}. 
At the same time, the results of modifying the material have a less obvious improvement on Non-referenced IQA because NR-IQA pays more attention to the beauty and clarity of the image \cite{NIMA_2018}. 

At last, our rendering results need further improvement because there are still obvious differences between our model and the real-world parking lot pictures visible to the naked eye.

\section{Conclusion}
Our research system has been successfully constructed. Although this is only a preliminary system, it is worthy of further improvement in almost every place. This will be the goal of our next stage of work.
In the future, we would like to extend this work to the applications of robotics \cite{lan2019evolutionary,lan2019simulated,lan2021learning} and target tracking \cite{xiang2016uav}.
\bibliographystyle{IEEEtran}
\bibliography{ref}

\begin{thebibliography}{10}
\providecommand{\url}[1]{#1}
\csname url@samestyle\endcsname
\providecommand{\newblock}{\relax}
\providecommand{\bibinfo}[2]{#2}
\providecommand{\BIBentrySTDinterwordspacing}{\spaceskip=0pt\relax}
\providecommand{\BIBentryALTinterwordstretchfactor}{4}
\providecommand{\BIBentryALTinterwordspacing}{\spaceskip=\fontdimen2\font plus
\BIBentryALTinterwordstretchfactor\fontdimen3\font minus
  \fontdimen4\font\relax}
\providecommand{\BIBforeignlanguage}[2]{{%
\expandafter\ifx\csname l@#1\endcsname\relax
\typeout{** WARNING: IEEEtran.bst: No hyphenation pattern has been}%
\typeout{** loaded for the language `#1'. Using the pattern for}%
\typeout{** the default language instead.}%
\else
\language=\csname l@#1\endcsname
\fi
#2}}
\providecommand{\BIBdecl}{\relax}
\BIBdecl

\bibitem{lan2016development1}
G.~Lan, Z.~Luo, and Q.~Hao, ``Development of a virtual reality teleconference
  system using distributed depth sensors,'' in \emph{2016 2nd IEEE
  International Conference on Computer and Communications (ICCC)}.\hskip 1em
  plus 0.5em minus 0.4em\relax IEEE, 2016, pp. 975--978.

\bibitem{lan2016development2}
G.~Lan, J.~Sun, C.~Li, Z.~Ou, Z.~Luo, J.~Liang, and Q.~Hao, ``Development of
  uav based virtual reality systems,'' in \emph{2016 IEEE International
  Conference on Multisensor Fusion and Integration for Intelligent Systems
  (MFI)}.\hskip 1em plus 0.5em minus 0.4em\relax IEEE, 2016, pp. 481--486.

\bibitem{CVRKD_2022}
G.~Y. et~al., ``Content-variant reference image quality assessment via
  knowledge distillation,'' \emph{The Thirty-Sixth AAAI Conference on
  Artificial Intelligence}, no.~9, 2022.

\bibitem{liu2022towards}
T.~Liu, G.~Lan, K.~A. Feenstra, Z.~Huang, and J.~Heringa, ``Towards a knowledge
  graph for pre-/probiotics and microbiota--gut--brain axis diseases,''
  \emph{Scientific Reports}, vol.~12, no.~1, p. 18977, 2022.

\bibitem{lan2022semantic}
G.~Lan, T.~Liu, X.~Wang, X.~Pan, and Z.~Huang, ``A semantic web technology
  index,'' \emph{Scientific reports}, vol.~12, no.~1, p. 3672, 2022.

\bibitem{WaDIQaM_2018}
S.~B. et~al., ``Deep neural networks for no-reference and full-reference image
  quality assessment,'' \emph{IEEE Transactions on Image Processing}, 2018.

\bibitem{NIMA_2018}
H.~Talebi and P.~Milanfar, ``Nima: Neural image assessment,'' \emph{IEEE
  Transactions on Image Processing}, 2018.

\bibitem{amann2013using}
J.~Amann, B.~Weber, and C.~A. W{\"u}thrich, ``Using image quality assessment to
  test rendering algorithms,'' \emph{International Conference in Central Europe
  on Computer Graphics and Visualization}, 2013.

\bibitem{lavou2014a}
G.~Lavoué and R.~Mantiuk, ``Quality assessment in computer graphics,'' in
  \emph{Visual Signal Quality Assessment}, 2014, p. 243–286.

\bibitem{mantiuk2011a}
\BIBentryALTinterwordspacing
R.~Mantiuk, ``\BIBforeignlanguage{et}{Hdr-vdp-2},'' in
  \emph{\BIBforeignlanguage{et}{ACM SIGGRAPH 2011}}, 2011, available at:.
  [Online]. Available: \url{https://doi.org/10.1145/1964921.1964935.}
\BIBentrySTDinterwordspacing

\bibitem{herzog2012a}
\BIBentryALTinterwordspacing
R.~Herzog, M.~Čadík, T.~O. Aydčin, K.~I. Kim, K.~Myszkowski, and H.-P.
  Seidel, ``\BIBforeignlanguage{en}{Norm: No-reference image quality metric for
  realistic image synthesis},'' \emph{\BIBforeignlanguage{en}{Computer Graphics
  Forum}}, vol.~31, no. 2pt3, p. 545–554, 2012. [Online]. Available:
  \url{https://doi.org/10.1111/j.1467-8659.2012.03055.x}
\BIBentrySTDinterwordspacing

\bibitem{lan2022vision}
G.~Lan, Y.~Wu, F.~Hu, and Q.~Hao, ``Vision-based human pose estimation via deep
  learning: A survey,'' \emph{IEEE Transactions on Human-Machine Systems},
  vol.~53, no.~1, pp. 253--268, 2023.

\bibitem{lan2022class}
G.~Lan, Z.~Gao, L.~Tong, and T.~Liu, ``Class binarization to neuroevolution for
  multiclass classification,'' \emph{Neural Computing and Applications},
  vol.~34, no.~22, pp. 19\,845--19\,862, 2022.

\bibitem{gao2021neat}
Z.~Gao and G.~Lan, ``A neat-based multiclass classification method with class
  binarization,'' in \emph{Proceedings of the genetic and evolutionary
  computation conference companion}, 2021, pp. 277--278.

\bibitem{GI}
T.~Ritschel, C.~Dachsbacher, T.~Grosch, and J.~Kautz, ``The state of the art in
  interactive global illumination,'' \emph{Computer Graphics Forum}, vol.~31,
  no.~1, p. 160–188, 2012.

\bibitem{epic_games_2020}
\BIBentryALTinterwordspacing
Epic-Games, ``Global illumination,'' 2020. [Online]. Available:
  \url{https://docs.unrealengine.com/4.26/en-US/RenderingAndGraphics/GlobalIllumination/}
\BIBentrySTDinterwordspacing

\bibitem{luksch_wimmer_schwarzler_2019}
C.~Luksch, M.~Wimmer, and M.~Schwärzler, ``Incrementally baked global
  illumination,'' \emph{Proceedings of the ACM SIGGRAPH Symposium on
  Interactive 3D Graphics and Games}, 2019.

\bibitem{botsch_2011}
M.~Botsch, \emph{Polygon Mesh Processing}.\hskip 1em plus 0.5em minus
  0.4em\relax A K Peters, 2011.

\bibitem{DisneyBRDF}
\BIBentryALTinterwordspacing
B.~Burley, ``Physically based shading at disney,'' 2012. [Online]. Available:
  \url{https://disneyanimation.com/publications/physically-based-shading-at-disney/}
\BIBentrySTDinterwordspacing

\bibitem{lan2019evolutionary}
G.~Lan, J.~Chen, and A.~Eiben, ``Evolutionary predator-prey robot systems: From
  simulation to real world,'' in \emph{Proceedings of the genetic and
  evolutionary computation conference companion}.\hskip 1em plus 0.5em minus
  0.4em\relax IEEE, 2019, pp. 123--124.

\bibitem{lan2019simulated}
G.~Lan, J.~Chen, and A.~E. Eiben, ``Simulated and real-world evolution of
  predator robots,'' in \emph{2019 IEEE Symposium Series on Computational
  Intelligence (SSCI)}.\hskip 1em plus 0.5em minus 0.4em\relax IEEE, 2019, pp.
  1974--1981.

\bibitem{lan2021learning}
G.~Lan, M.~van Hooft, M.~De~Carlo, and J.~M. Tomczak, ``Learning locomotion
  skills in evolvable robots,'' \emph{Neurocomputing}, vol. 452, pp. 294--306,
  2021.

\bibitem{xiang2016uav}
T.~Xiang, F.~Jiang, G.~Lan, J.~Sun, G.~Liu, Q.~Hao, and C.~Wang, ``Uav based
  target tracking and recognition,'' in \emph{2016 IEEE international
  conference on multisensor fusion and integration for intelligent systems
  (MFI)}.\hskip 1em plus 0.5em minus 0.4em\relax IEEE, 2016, pp. 400--405.

\end{thebibliography}

\end{document}